# A Comprehensive Literature Review on Sweet Orange Leaf Diseases

Yousuf Rayhan Emon

Teaching Assistant

Department of Computer Science and Engineering

Daffodil International University, Dhaka, Bangladesh

yousuf15-3220@diu.edu.bd

Md Golam Rabbani

Research Student

Department of Computer Science and Engineering

Daffodil International University, Dhaka, Bangladesh

golam15-4184@diu.edu.bd

Dr. Md. Taimur Ahad

Associate Professor

Department of Computer Science and Engineering

Daffodil International University, Dhaka, Bangladesh

taimurahad.cse@diu.edu.bd

Faruk Ahmed

Research Student

Department of Computer Science and Engineering

Daffodil International University, Dhaka, Bangladesh

golam15-4184@diu.edu.bd

***Abstract:*** *Sweet orange leaf diseases are significant to agricultural productivity. Leaf diseases impact fruit quality in the citrus industry. The apparition of machine learning makes the development of disease finder. Early detection and diagnosis are necessary for leaf management. Sweet orange leaf disease-predicting automated systems have already been developed using different image-processing techniques. This comprehensive literature review is systematically based on leaf disease and machine learning methodologies applied to the detection of damaged leaves via image classification. The benefits and limitations of different machine learning models, including Vision Transformer (ViT), Neural Network (CNN), CNN with SoftMax and RBF SVM, Hybrid CNN-SVM, HLB-ConvMLP, EfficientNet-b0, YOLOv5, YOLOv7, Convolutional, Deep CNN. These machine learning models tested on various datasets and detected the disease. This comprehensive review study related to leaf disease compares the performance of the models; those models' accuracy, precision, recall, etc., were used in the subsisting studies.*

## Introduction:

Deep learning, a branch of artificial intelligence, has shown great promise in various fields, including the recognition of diseases in citrus fruits and leaves. This method, combining deep learning with image processing, helps farmers tell apart healthy crops from those that are infected. Specifically, we use convolutional neural networks (CNN) on a cloud-based Platform as a Service (PaaS) to identify diseased and healthy citrus leaf images (Lanjewar & Parab, 2023). Additionally, a specialized CNN model, fine-tuned and pre-trained, has been developed to detect black spot disease and determine the ripeness of orange fruit (Momeny et al. 2022). The growth of machine learning technologies has opened new doors in fighting these diseases (Ahmed et al., 2023). It's especially crucial to detect citrus diseases early and accurately since they are prone to pest attacks, which can greatly harm the crops (Lee et al 2022).

Deep Learning models, noted for their multiple layers and vast numbers of parameters, have been successfully applied in agriculture for tasks like disease detection and fruit counting. However, these models require significant processing power, which can be a challenge for mobile devices. This is further complicated by often poor internet connectivity in fields, making cloud computing difficult and shifting the focus towards edge computing solutions. Deep learning has also proven to be a valuable tool in the manufacturing industry, particularly in classifying food products, such as identifying sweet oranges in factories (Salaiwarakul & Mungklachaiya, 2023).

Huanglongbing (HLB), a severe threat to citrus production, has caused considerable economic damage globally. CNN-based computer-vision systems have shown an accuracy in detecting HLB (Gómez-Flores et al 2022). To effectively manage these diseases, precise and effective identification and classification techniques are needed. For classifying different lemon diseases, this paper proposes a hybrid model combining Support Vector Machines (SVM) and CNN (Gupta et al. 2023, May). The high accuracy of CNNs in detecting and categorizing leaf diseases indicates their promise in the field of plant disease detection, potentially having a significant impact on real-time agricultural systems (Ahad et al., 2023).

The early detection of bruising in fruits, such as lemons, is important to prevent low-quality produce from reaching the market. A study has focused on the early identification of bruised lemons using 3D-CNNs and a local spectral-spatial hyperspectral imaging technique, which analyzes adjacent pixel information in both frequency and spatial domains (Pourdarbani et al. 2023). Early diagnosis and detection are crucial for improving the management of leaf diseases in plants (Mustofa et al., 2023; Bhowmik et al., 2023). Manually inspecting fruits and leaves to classify disease symptoms is laborious and time-consuming. There's a need for a computer vision system that autonomously classifies fruits and leaves, speeding up management in fields (Yadav et al., 2023, June). Diseases in leaves and fruits cause significant harm to crops, reducing both the quality and quantity of the produce. Timely identification of these infections is key to enhancing plant productivity, yet it remains a challenging task (Lanjewar & Parab, 2023). Bruises on fruits often signify cell damage, which can lead to decreased protection against oxygen, causing the breakdown of cell walls and membranes. This results in oxidation, producing an unwanted brown coloration due to enzyme reactions (Shastri et al., 2023).

The early detection and accurate classification of citrus canker disease are essential for effective disease management and control. This research suggests a hybrid model that combines CNN and SVM for the multi-classification of lemon citrus canker diseases (Kukreja et al 2023, May). The development of technology, especially in image processing and machine learning, offers opportunities for early detection and classification of plant diseases (Çetiner 2022). HLB, a major threat to citrus production, has led to significant economic losses worldwide. It's important that the detection system differentiates HLB from other citrus abnormalities for effective treatment. The primary detection method for HLB's causal pathogen is the expensive quantitative real-time polymerase chain reaction (qPCR) test, making it difficult to gather large datasets for training CNN-based systems (Gómez-Flores et al 2022). Citrus black spot (CBS) is a fungal disease that limits market access for affected fruit, causing lesions on fruit surfaces and potentially leading to premature fruit drops, which reduce yield. Leaf symptoms for CBS are rare, and the fungus reproduces in leaf litter. Citrus canker, caused by the bacterium Xanthomonas citri subsp. citri, leads to economic losses due to fruit drops and blemishes (Yadav et al. 2023, June). Citrus fruit diseases are a major cause of significant declines in citrus fruit yield, highlighting the importance of developing automated detection systems for citrus plant diseases (Lanjewar, & Parab 2023).

Orange, a globally grown citrus fruit, is popular among health-conscious individuals due to its nutritional value. Classifying oranges is crucial for quality control, sorting, and grading in the food industry. However, farm-based disease prediction is not fully utilizing available technology (Garg et al. 2023, May). Diseases and varying ripeness levels in fruits affect their marketability, economic value, and increase crop waste (Momeny et al. 2022). The agricultural industry, particularly in lemon cultivation and production, faces challenges from diseases such as Lemon Scab, Septoria Spot, Sooty Mould, Armillaria, and Huanglongbing, which affect both the yield and quality of the fruit, as well as the growth of lemon trees (Gupta et al. 2023, May). Accurate and timely detection of diseases in citrus crops is critical for effective crop management and preventing yield loss. Traditional disease detection methods, like visual inspection, are often slow and error-prone (Shastri et al. 2023). Lemon citrus canker disease, a bacterial infection, causes significant economic losses in citrus production worldwide (Kukreja et al 2023, May). Plant diseases, which disrupt the food supply, lead to reduced agricultural yield and production, causing significant economic losses. Citrus plants, being widely grown and economically important, suffer from yield and quality losses due to pests and various diseases (Çetiner 2022). Lemon disease detection has been a focus of research due to increasing demand and prevalence of diseases in the crop (Sharma & Kukreja 2022, March).

Artificial intelligence is increasingly used for quality assessment in agriculture. Deep learning and machine learning methods process images of agricultural products to evaluate their quality, classifying them according to specific standards (Dümen et al. 2023). The occurrence of citrus Huanglongbing (HLB) has caused great losses in the citrus industry. Research is focused on controlling this disease using hyperspectral imaging technology and lightweight neural network technology. This study aims to improve the accuracy of HLB detection and its deployment at the end-edge (Peng et al 2022, July). Detecting and providing early warnings for citrus psyllids is crucial as they are the primary vectors of citrus huanglongbing (Lyu et al 2023). HLB is a serious worldwide citrus disease. Rapid, onsite, and accurate field detection of HLB has been challenging. A novel HLB detection method combines headspace solid-phase microextraction with portable gas chromatography–mass spectrometry (PGC-MS) for detecting volatile metabolites in citrus leaves (Liu et al 2023). Quality assessment in agriculture is key to production efficiency and marketability, improving industry standards, increasing sales, and reducing crop loss (Dümen et al. 2023). Plant diseases cause a major reduction in agricultural output, leading to severe economic losses and an unstable food supply. The citrus plant, an important crop worldwide, is susceptible to factors like climate change, pests, and diseases, resulting in reduced yield and quality (Lee et al 2022). HLB, or citrus greening disease, has variable symptoms that are hard to diagnose, resulting in low efficiency (Qiu et al. 2022). Classifying oranges based on sweetness usually requires manual inspection and tasting, but deep learning can automate this process for faster and more consistent results (Salaiwarakul & Mungklachaiya, 2023).

## Process of Literature Review:

This study is related to a comprehensive literature review for detection, feature extraction, feature extraction using traditional machine learning techniques for the classification, autoencoders (AEs), CNN (Convolutional Neural Network), and hybrid models for deep learning techniques included in this study. Most importantly, the research that experimented using a proper research methodology without providing experimental research was included in the literature review.

### Article Selection:

We utilized a detailed, three-phase selection process to choose articles for our final review, adhering strictly to predefined inclusion and exclusion standards. Initially, we eliminated any repetitive entries that resulted from database searches. Then, we conducted an initial sift through the titles, followed by an evaluation of the abstracts and, ultimately, a thorough examination of the complete texts to verify their relevance. The principal researcher was responsible for the entire selection procedure.

For an article to be included in our review, it needed to meet several criteria:

- It must be an original study published in a peer-reviewed journal, with the full text accessible through our university's resources.
- It should discuss the usage of plant leaves in its research and include relevant imagery.
- The publication language must be English.
- It should focus on the application of machine learning techniques to detect diseases in plant leaves.
- We concentrated on articles released between 2018 and 2023 to emphasize recent advancements in deep learning approaches.

On the other hand, we excluded articles that were:

- Entire books or individual book chapters.
- Brief reports or individual case studies.
- Studies with vague data descriptions.
- Research without any validation of the methods or findings.

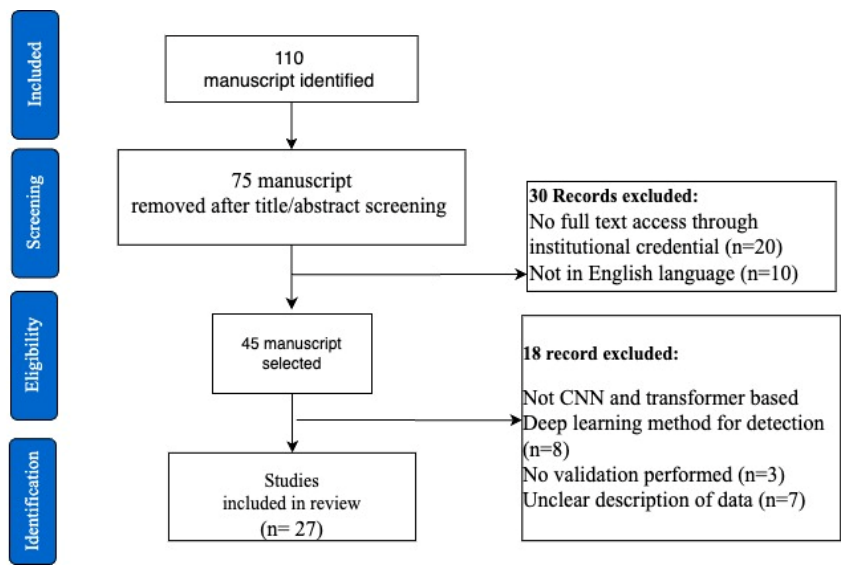

**Fig : Article Selection Process**

# Literature Review:

The literature review consists of following sections:

**Convolutional Neural Networks (CNN) in Sweet Orange Disease Detection**

Recent research in sweet orange leaf disease detection has seen a pronounced emphasis on the use of Convolutional Neural Networks (CNN). Gómez-Flores et al. (2022) contributed to this domain by fine-tuning pre-trained CNNs to differentiate between HLB and other citrus abnormalities. Similarly, Yadav et al. (2023) demonstrated the proficiency of CNNs in classifying the surface conditions of Valencia oranges and Furr mandarin leaves. A custom CNN, complemented by SVM, was employed to achieve a high degree of accuracy. Khattak et al. (2021) advanced this approach by developing a CNN model that excelled in identifying common citrus diseases, validating its effectiveness as a farmer's aid.

Lanjewar & Parab (2023) enriched this pool of knowledge by classifying healthy and diseased citrus leaves using deep CNN models, with data augmentation playing a pivotal role in enhancing model performance. Furthermore, Gupta et al. (2023) and Pourdarbani et al. (2023) harnessed CNNs for detecting diseases in lemon leaves and classifying bruising in lemons, respectively, showcasing the adaptability of CNNs in various citrus health monitoring scenarios.

**Transfer Learning for Enhanced Model Efficacy in Sweet Orange Disease Detection**

A subset of studies has focused on transfer learning to improve the detection and classification of sweet orange diseases. Gómez-Flores et al. (2022) emphasized the significance of the number of trainable parameters over network depth, with VGG19 achieving perfect sensitivity in HLB detection. This insight has bolstered the understanding of how transfer learning can be tailored to suit the needs of small-sized datasets prevalent in agricultural research.

**Ensemble and Hybrid Models for Sweet Orange Disease Detection**

The fusion of different machine learning techniques has resulted in the formation of ensemble and hybrid models, aiming to leverage the strengths of various algorithms. Garg et al. (2023) unveiled a hybrid model combining CNN and SVM, achieving notable accuracy in classifying six common disorders in oranges. The model's sensitivity analysis revealed critical features for disease categorization, underscoring the value of hybrid models in the precise diagnosis of citrus ailments. Momeny et al. (2022) introduced an innovative 'learning-to-augment' strategy, sidestepping traditional data augmentation methods to enhance the robustness of the model. Their approach, which involved fine-tuning pre-trained models with an augmented dataset, resulted in a substantial improvement in accuracy for black spot disease detection.

**Advanced Techniques and Novel Implementations for Disease Detection**

In search of advanced solutions, several researchers have implemented novel techniques for sweet orange disease detection. Shastri et al. (2023) introduced an Enhanced-CNN model that, through meticulous layer investigation and image processing, achieved high marks in disease detection and classification, outperforming previous methods. Kukreja et al. (2023) trained a model on citrus canker severity levels, achieving higher accuracy than existing models, which bodes well for practical disease management applications.

Çetiner (2022) designed a unique CNN architecture for detecting blackspot, canker, and greening diseases, achieving impressive performance metrics. This model serves as a decision-support tool, enhancing farmers' ability to recognize and classify citrus diseases.

**Emerging Technologies and Future Directions**

Exploring emerging technologies, Dümen et al. (2023) and Peng et al. (2022) ventured into the realm of data augmentation and hybrid modeling, respectively. Dümen et al. showcased the ViT method's unparalleled accuracy in lemon classification, while Peng et al. presented a hybrid CNN-MLP model that exhibited high accuracy with a low parameter footprint, ideal for end-edge deployment. Lee et al. (2022) and da Silva et al. (2023) employed transfer learning models to develop web applications and mobile-edge AI solutions for citrus pest diagnosis, enabling the use of AI algorithms in practical, real-world scenarios. Lyu et al. (2023) proposed a lightweight detection model for citrus psyllids, optimizing hyperparameters with the black widow optimization algorithm, which showed promising results for practical applications in natural environments.

Qiu et al. (2022) created an automatic HLB identification model, leading to the development of an app that assists farmers in rapid HLB detection, while Apacionado & Ahamed (2023) focused on nighttime detection of sooty molds, showcasing the potential for 24/7 disease monitoring. Aswini & Vijayakumaran (2023) overcame the challenge of subjective diagnosis by employing YOLOv7, achieving high accuracy across diverse datasets, which could greatly assist in the management of citrus greening disease.

## Table-1: Research Matrix:

| Author | Model | Accuracy | Contribution |
|---|---|---|---|
| Gómez-Flores et al (2022) | CNN | 95% | Pre-trained CNNs are fine-tuned to distinguish HLB. |
| Yadav et al. (2023, June) | CNN with SoftMax and RBF SVM | 89.8% and 92.1% | Detection and management of groves infected by CBS. |
| Khattak et al. (2021). | CNN | 94.55% | Automatic detection of citrus fruit and leaves diseases. |
| Lanjewar, & Parab (2023) | CNN | 98% | Citrus leaf disease detection using PaaS cloud on mobile. |
| Garg et al. (2023, May). | Hybrid CNN-SVM | 88.13734% | Efficient Detection and Classification of Orange Diseases. |
| Momeny et al. (2022). | CNN | 99.5% | Detection of citrus black spot disease and ripeness. |
| Gupta et al. (2023, May). | Hybrid CNN-SVM | 89.6% | Lemon Diseases Detection and Classification using Hybrid CNN-SVM Model. |
| Pourdarbani et al. (2023) | CNN | 90.47% | Examination of Lemon Bruising Using Different CNN-Based. |
| Shastri et al. (2023) | E-CNN | recognition 98%, classification 99%. | Detection of Citrus Fruits and Leaves Diseases Using Enhanced Convolutional Neural Network. |

| Author | Model | Accuracy | Description |
|---|---|---|---|
| Kukreja et al (2023, May) | CNN | 94.03% | Efficient and Accurate Diagnosis of Lemon Citrus Canker Disease |
| Çetiner (2022) | CNN | 96% | Citrus disease detection and classification based on convolution deep neural network. |
| Sharma & Kukreja (2022, March) | CNN | 98.43% | network (CLTN) model for Lemon Citrus Canker Disease Multi-classification. |
| Dümen et al. (2023) | ViT | 99.84% | Achieving High Accuracy in Lemon Quality Classification. |
| Peng et al (2022, July) | HLB-ConvMLP | 95.18% | HLB-ConvMLP-Rapid Identification of Citrus Leaf Diseases. |
| Lee et al (2022) | EfficientNet-b0 | 97% | Automatic Classification Service System for Citrus Pest Recognition Based on Deep Learning. |
| da Silva et al. (2023) | Yolo | 98% | Using Mobile Edge AI to Detect and Map Diseases in Citrus Orchards. |
| Lyu et al (2023) | Yolo | 97.18% | Proposed a detection model for citrus psyllid based on spatial channel interaction. |
| Qiu et al. (2022) | Yolo | micro F1-score of 84.64% and 85.84% | An automatic identification system for citrus greening disease. |
| Apacionado & Ahamed (2023) | YOLOv5m, YOLOv7, and CenterNet | mAP: YOLOv7: 74.4% | Proposed a cost-effective nighttime sooty mold detection method for citrus canopies using a |

| | | YOLOv5m: 72% CenterNet: 70.3% | home surveillance camera and YOLOv7. |
|---|---|---|---|
| Aswini & Vijayakumaran (2023) | YOLOV7 | F1 score: 92%. | Demonstrated YOLOv7's robustness in diagnosing Huanglongbing in citrus plants through image analysis. |
| Song et al. (2020) | YOLOv4 | mAP: 95.40% | Introduced a YOLO-based automated system for swiftly identifying Citrus Canker and Citrus Greening from leaf images. |
| Salaiwarakul & Mungklachaiya (2023) | VGG-16 | 83.57% | A Simple Deep Learning Model for Classifying Oranges in Embedded Manufacturing. |
| Liu et al (2023). | PGC–MS | 93.3% | Affected Sweet Orange Juices Using a Novel Combined Strategy of Untargeted Metabolomics and Machine Learning. |

**Inference from current research studies:**

According to the research matrix (see Table 1), there has been a recent upswing in citrus leaf disease detection research, fueled by the integration of cutting-edge machine learning models. In the domain of citrus disease detection, a range of research emphasizes the effectiveness of advanced deep learning models, notably CNN architectures and hybrid methods, in accurately identifying and categorizing diseases impacting citrus fruits and leaves. Studies by Gomez-Flores et al. (2022) and Yadav et al. (2023) particularly highlight the crucial role played by CNN architectures in pinpointing ailments like Huanglongbing (HLB) and classifying diverse citrus abnormalities. Their findings emphasize that the number of trainable parameters significantly influences detection precision, showcasing VGG19's exceptional sensitivity due to its ability to effectively transfer learned traits. Moreover, research from Khattak et al. (2021) and Shastri et al. (2023) underscores the superior performance of CNN models, positioning them as valuable tools aiding farmers and inspectors in disease identification and agricultural management. Collectively, these studies underscore how CNN architectures and hybrid models are reshaping disease detection and control strategies in the citrus industry, potentially elevating both crop yield and quality. Moreover, the utilization of augmentation methodologies as exemplified in the works of Lanjewar & Parab (2023) and the innovative strategies employed by Momeny et al. (2022) denote significant strides in augmenting datasets and monitoring diseases. Additionally, the propositions set forth by Garg et al. (2023) and Gupta et al. (2023), amalgamating CNN with supplementary algorithms such as SVM, manifest heightened precision in the classification of citrus disorders and meticulous disease identification. These cumulative initiatives underscore the burgeoning potential inherent in deep learning paradigms and hybrid models, promising to reform the practices surrounding disease diagnosis, classification, and management. This trajectory holds the potential to redefine the paradigm of citrus cultivation and strategies for disease containment and mitigation.

Furthermore, the reviewed studies focused on leveraging cutting-edge AI and deep learning techniques to address the detection, identification, and management of diseases affecting citrus crops. da Silva et al. (2023) put forth a system designed with minimal computational demands, evaluating YOLO and Faster R-CNN for fruit detection, alongside MobileNetV2, EfficientNetV2-B0, and NASNet-Mobile for classification, providing a comparative analysis of their performance and computational efficiency. Lyu et al. (2023) introduced a tailored YOLO-SCL model optimized for detecting citrus psyllids, boasting high accuracy, reduced parameter count, and suitability for natural settings. Qiu et al. (2022) crafted YOLOv5l models specialized in identifying Huanglongbing (HLB) symptoms, notably highlighting the superior recognition precision of the Yolov5l-HLB2 model, potentially serving as an initial screening tool. Apacionado & Ahamed (2023) explored nocturnal disease detection using specialized cameras and found YOLOv7 particularly proficient in identifying sooty molds on citrus canopies at night. Aswini & Vijayakumaran (2023) addressed the complexities of citrus greening disease diagnosis employing YOLOv7, demonstrating remarkable accuracy across diverse datasets. Finally, Song et al. (2020) introduced an automated system utilizing YOLO for disease identification in citrus leaf images and videos, showcasing the model's potential in annotating disease instances. These studies collectively underscore the efficacy of AI-driven models, particularly variants of YOLO, as promising tools for disease detection, classification, and ongoing monitoring within citrus cultivation.

The collective body of research presented an array of advanced technological applications within citrus agriculture and disease identification. Dümen et al. (2023) employed innovative methods, augmenting lemon datasets and utilizing transformer-based models, notably ViT, to achieve the most precise citrus classification, showcasing the promise of transformer models and data augmentation in accurate citrus quality assessment. Meanwhile, Peng et al. (2022) introduced HLB-ConvMLP, a hybrid model merging convolutional neural networks and multilayer perceptrons, demonstrating enhanced accuracy, reduced complexity, and faster inference rates in detecting Huanglongbing (HLB) compared to conventional CNNs, underlining its potential for real-time disease identification. Lee et al. (2022) compiled a comprehensive dataset and utilized transfer learning, achieving notable accuracy in identifying citrus pests, leading to the creation of a web-based tool for diagnosing citrus-related issues. This highlights the tangible benefits of disease identification and classification in citrus farming. On the other hand, Salaiwarakul & Mungklachaiya (2023) introduced a deep learning model for determining the sweetness of oranges, revealing that strategies like dropout and image augmentation significantly boosted the performance of CNN models such as ResNet-50 and VGG-16, underscoring the importance of these approaches in refining classification accuracy. Finally, Liu et al. (2023) pioneered a pioneering method for on-site HLB detection by analyzing volatile metabolites in citrus leaves through a novel application of the random forest algorithm. This approach showcased remarkable accuracy and speed in identifying healthy and HLB-affected trees, offering a prospective avenue for rapid on-field diagnosis of HLB. In summary, these studies collectively emphasize the transformative capacity of sophisticated technologies such as deep learning, hybrid models, and inventive detection techniques in reshaping disease diagnosis, quality evaluation, and surveillance methodologies within the citrus agricultural domain.

## Limitations:

The primary purpose of this study is to comprehend the current landscape in sweet orange disease detection using deep learning techniques. This research aims to the sweet orange leaf disease related to all the paper collections and their comprehensive model using and their accuracy and which was the best for finding the best result for sweet oranges. This research direction is closely aligned with the object of harnessing technology to improve disease management strategies and promote sweet orange production.

It is noticeable that existing literature publications commonly train using the same model many times. Most of the papers are directed using one dataset. They're detecting disease in sweet orange training freely accessible dataset and evaluating a model for detection. However, it is not an idea for the other datasets to perform using the model. In addition, it is important to recognize the availability of large amounts of image data and the effectiveness of deep learning methods in disease classification. Gaining extensive and several dataset images accurately labeled with examples of various diseases poses a significant challenge.

Building datasets like these requires a complex and time-consuming process that creates difficulty. Sweet orange leaf diseases come in many forms and are affected by the growth stages and how they appear, which makes reliable datasets challenging. Developing deep learning models effectively demands various and high-quality training data, which face difficult challenges. Researchers who focus on sweet orange leaf disease gather a lot of pictures with accurate labels to identify different diseases in every image. This requires knowledge and muscular time commitment. As a result, a lot of researchers have decided to use already existing datasets.

**Direction for Future Research:**

The future research directions for improving the accuracy and efficiency of sweet orange leaf disease detection using Machine learning include reinforcement learning, hybrid machine learning, and case-based reasoning. Reinforcement learning can be used to develop a machine-learning model that can automatically identify diseased sweet orange leaves without the need for human intervention. Hybrid machine learning can combine several models with a traditional machine learning algorithm to improve the accuracy of sweet orange leaf disease detection. Case-based reasoning can be used to develop a machine-learning model that can learn from the mistakes of previous models and improve its performance over time.

In addition to these novel research directions, several other challenges must be addressed to improve the accuracy and efficiency of sweet orange leaf disease detection using machine learning models. These challenges include data scarcity, environmental variability, and real-time detection. Despite these challenges, machine learning models are promising technology for sweet orange leaf disease detection. By continuing to research novel techniques and address the existing challenges, it is possible to develop some machine learning models that can accurately and efficiently detect sweet orange leaf diseases.

## Conclusion:

This study provides a comprehensive review of sweet orange leaf disease detection and investigates various approaches and methodologies for the early identification and treatment of plant disease. Neural Network (CNN) has surfaced as a potentially useful technology for classifying and detecting diseases of the sweet orange leaf in synchronism with other models like Transformers (ViT), approaches based on Yolo, and hybrid models. Researchers have evaluated the advantages and disadvantages of these models and conducted evaluations using metrics like recall, accuracy, and precision. While ViT, CNN models, and other innovative techniques show strength, it's critical to evaluate their practicality, especially with datasets of sweet orange leaves. Additionally, models, including vision transformers and ensembled models, have shown their efficiency, while advanced CNN models and YOLO-based models have proved effective in recognizing various sweet orange leaf diseases. Together, these studies provide valuable information about the development of sweet orange leaf disease detection, with a focus on improving farming methods and addressing the unique difficulties that farmers face in their real-life situations.